\documentclass[runningheads]{llncs}
\usepackage{graphicx}
\usepackage{wrapfig}
\usepackage{booktabs}
\usepackage{algorithm}
\usepackage{algpseudocode}
\usepackage{amsmath}
\usepackage{amssymb} % for Greek symbols
\usepackage{subcaption}
\usepackage{setspace}
\usepackage{ragged2e}
\begin{document}
\title{Fact-Checking Generative AI: \\Ontology-Driven Biological Graphs for Disease-Gene Link Verification}

\author{Ahmed Abdeen Hamed\inst{1} \and
Alessandro Crimi\inst{1}  \and
Byung Suk Lee\inst{2} \and 
Magdalena M. Misiak \inst{3}}
\authorrunning{Ahmed Abdeen Hamed et al.}
% First names are abbreviated in the running head.
% If there are more than two authors, 'et al.' is used.
%
\institute{Sano Centre for Computational Medicine, Cracow, Poland \and
\email{a.hamed@sanoscience.org}\\ \and 
University of Vermont, Department of Computer Science, Burlington VT, USA \and 
Department of Physiology and Biophysics, Howard University Heidelberg, Germany \and 
Department of Physiology and Biophysics, Washington DC, USA}

\maketitle              % typeset the header of the contribution
\begin{abstract}
Since the launch of various generative AI tools, scientists have been striving to evaluate their capabilities and contents, in the hope of establishing trust in their generative abilities. Regulations and guidelines are emerging to verify generated contents and identify novel uses. we aspire to demonstrate how ChatGPT claims are checked computationally using the rigor of network models. We aim to achieve fact-checking of the knowledge embedded in biological graphs that were contrived from ChatGPT contents at the aggregate level. We adopted a biological networks approach that enables the systematic interrogation of ChatGPT's linked entities. We designed an ontology-driven fact-checking algorithm that compares biological graphs constructed from approximately 200,000 PubMed abstracts with counterparts constructed from a dataset generated using the ChatGPT-3.5 Turbo model. In 10-samples of 250 randomly selected records a ChatGPT dataset of 1000 ``simulated'' articles , the fact-checking link accuracy ranged from 70\% to 86\%. This study demonstrated high accuracy of aggregate disease-gene links relationships found in ChatGPT-generated texts. 

% \keywords{First keyword  \and Second keyword \and Another keyword.}
\end{abstract}

\section{Introduction}
The rise of new generative AI technologies holds both potential and concerns. Particularly, the emergence of ChatGPT \cite{chatgpt} caused scientists to raise various concerns related to the capabilities and the inauthentic contents of such tools. Van Dis et al. identified five key priorities aimed at educating the general public about the potential of ChatGPT and formulating an effective response to this transformative AI tool. Among the five guidelines, fact-checking and human verification of ChatGPT contents were highlighted \cite{van2023chatgpt}. Inspired by such guidelines, here we present our work on computational fact-checking of biological networks we constructed from ChatGPT-generated content. The utilization of biological ontology (i.e., Disease Ontology, Gene Ontology, Gene Ontology Annotations) give credibility to the biological terms that make up the nodes of the graphs. Using biological entities from ontology to extract and construct biological graphs from the biomedical literature offers trustworthy ground truth. Using network models and algorithms offer the rigor needed to perform fact-checking at the aggregate level. This study assumes a closed-world assumption \cite{Przymusinski1989,v34i03,Torralba2011,Minker1982}, which sets the fact-checking scope within the knowledge embedded in the literature dataset and not beyond.

Knowledge graphs have been instrumental in advancing fact-checking methodologies, enabling structured and nuanced analyses of claims and assertions. For example, Tchechmedjiev et al. introduced ClaimsKG, a comprehensive knowledge graph that houses fact-checked claims, allowing informed queries on truth values and related aspects \cite{tchechmedjiev2019claimskg}. Vedula and Parthasarathy's work stood out by introducing FACE-KEG, a knowledge graph tailored to expound whether a statement is true or false, addressing the transparency gap in fact-checking \cite{vedula2021face}. Lin et al. made strides with ontology-based subgraph patterns, constructing graph fact-checking rules that integrate intricate patterns, capturing both topological and ontological constraints \cite{lin2018discovering,lin2018fact,lin2019discovering}. Notably, Ciampaglia et al. laid a foundation for fact-checking by leveraging knowledge graphs to scrutinize claims, drawing from reliable sources like Wikipedia \cite{ciampaglia2015computational}.

Wang et al. harnessed entity category information, using prototype-based learning to enhance verification accuracy and reasoning capabilities in knowledge graph-based fact-checking, marking a significant advancement in this domain \cite{Wang2020}. Khandelwal et al.'s approach encompassed structured and unstructured data from knowledge graphs to address the challenge of evaluating facts amidst growing data and misinformation \cite{Khandelwal2020}. Orthlieb et al.'s attention-based path ranking model exhibited promise in automating fact-checking through knowledge graphs, emphasizing interpretability and competitive results \cite{Orthlieb2021}. Another notable contribution came from Shi, who introduced ProjE, a neural network model that improved the completion of knowledge graphs and the accuracy of fact-checking \cite{Shi2017}.

Recent advancements further underpin the significance of knowledge graphs in fact-checking. The approach of Wang et al. leveraged category hierarchy and attribute relationships, showcasing the potential of knowledge structure in fact verification \cite{Wang2022}. Amidst the COVID-19 outbreak, Mengoni's extended knowledge graph enabled enhanced claim validation through leveraging existing fact-checking reports \cite{Mengoni2022}. Kim introduced weighted logical rules mining and evidential path identification in knowledge graphs, enhancing computational fact-checking \cite{Kim2020,Kim2021}. Zhu et al. designed a knowledge-enhanced fact-checking system, tapping into both unstructured document knowledge bases and structured graphs to robustly identify misinformation \cite{Zhu2021}.

% Approaches incorporating knowledge graphs have emerged as vital pillars in computational fact-checking. Lin et al.'s diverse methods, encompassing ontological knowledge graphs and utilizing discriminant subgraph structures, offered comprehensive solutions \cite{Lin2017,Lin2018,Lin2019}. Shahi's contribution was particularly timely, introducing FakeKG, a knowledge graph-based approach aimed at countering misinformation during crises and elections \cite{Shahi2023}. Qudus et al.'s hybrid fact-checking approach, encompassing multiple categories, demonstrated superior performance in ensemble learning \cite{Qudus2022}. Meanwhile, Shiralkar et al.'s network-flow-based knowledge graph method addressed the challenge posed by the abundance of online information \cite{Shiralkar2017}.

% Liu et al.'s KompaRe introduced a system for comparative reasoning across knowledge graphs, broadening reasoning capabilities beyond conventional methods \cite{Liu2021}. Luo et al.'s work emphasized the significance of fact-checking in detecting fake news, highlighting methodologies like Knowledge Linker, PredPath, and Knowledge Stream \cite{Luo2021}. Shi's approach to fact-checking in knowledge graphs, treated as a link-prediction problem, exhibited promising outcomes \cite{Shi2016}. These contributions collectively illustrate the dynamic landscape of knowledge graph-based fact-checking and its evolving potential to enhance information accuracy and credibility.

\section{Methods}\label{sec11}
In this section, we present a comprehensive methodology for constructing a reliable knowledge framework to assess the quality of content generated using ChatGPT. Our approach is centered around the utilization of biological graphs as rigorous models that offer quantitative analysis of objective outcomes. Graphs as a tool is also being investigated for the advancement of Large Language Models (LLMs) \cite{pan2023unifying} and ChatGPT technologies \cite{yang2023chatgpt}.

The proposed approach %integrates five key aspects 
consists of six key steps which as a whole contribute to verification of the authenticity and accuracy of AI-generated biomedical text: (1) ChatGPT prompt-engineering and simulated-articles generation, (2) partial-match ontology term chunking to increase the recall of term matching, (3) ontology feature extraction, where partial terms are used as the means to feature identification in the literature and ChatGPT text, (4) proximity-based biological graphs construction for capturing the strongest links among the biological terms, (5) biological graph topological analysis, by analyzing the structural properties of each type and comparing them accordingly, and (6) algorithmic fact-checking to assert the facts. 

% Figure \ref{fig:graphical-abstract} shows a graphical abstract of our approach in the various steps.  
% \begin{figure}
%   \includegraphics[scale=0.2]{Graphical-Abstract-GPT.png}
%   \caption{A graphical abstract that demonstrates the various computational steps of this study.}
%   \label{fig:graphical-abstract}
% \end{figure}

\subsection{Prompt-Engineering ChatGPT for Simulated-Articles Generation }
Using the ChatGPT APIs, we engineered a prompt that has two roles: (1) the system role which is to command the ChatGPT engine to generate biomedical abstracts and (2) a user role which is to command ChatGPT explicitly to perform the task shown in Algorithm \ref{algo:prompt-eng}, repeating it as needed until a dataset of the desired size is produced.

\begin{algorithm}
\caption{ChatGPT Prompt Engineering for Article Generation}
\label{algo:prompt-eng}
\begin{algorithmic}[1]
    \Require The number $n$ of simulated articles.
    \Require The number $w$ of words in each article.
    \State Generate a list of $n$ simulated PubMed-style abstracts.
    \State For each abstract containing three fields: GPT-ID, Title, and Abstract, make it $w$ words.
    \State Make the GPT-ID random, containing at most five letters and numbers.
    \State Return the abstracts in a valid JSON format as an array of JSON records.
    \State Investigate the biology of human disease-gene associations.
    \State Provide details related to diseases, genes, cells, organisms, and any FDA-approved drugs, and state any relationships.
\end{algorithmic}
\end{algorithm}

\subsection{Feature Extraction and Biological Graph Construction}
Ontology terms are inherently detailed and lengthy. In biomedical literature, these long names are frequently abbreviated for convenience. For instance, the term ``female breast cancer'' is often referred to as ``breast cancer'' in the text. Importantly, we maintained a connection between these bigrams and their corresponding original term IDs in the ontology, while also tracking their positions. Constructing the knowledge graphs required the following steps: (1) feature extraction using the diseases and gene ontology and (2) establishing the links among the terms extracted. The process of ontology feature extraction from text records is as follow: (1) it takes as in put a collection of abstract texts and an ontology containing terms, (2) reads each textual record in the collection to identifies mentions of ontology terms (and related bigrams) within the text, (3) checks if the term appears in the text. If the term is a single word, the algorithm records the position of the match, the term itself, and other relevant information. For terms with more than one word, the algorithm generates bigrams (pairs of adjacent words) and checks for their presence in the text. If found, it records the position, term, bigram, and additional information. The process terminates by producing a set of matches for each record, indicating where ontology terms and bigrams were found within the text. Concretely, we constructed two different undirected but weighted graphs of disease and genes nodes. The first type one was constructed publication-driven  from the mentions of disease and genes occurring in a dataset of biomedical abstracts extracted from PubMed Central \cite{PMC}. A disease and a gene are connected if they occur in the same abstract. Then the link is weighted with the distance among the terms. Both gene and diseases names are ontology terms from the Human Disease Ontology (DOID) \cite{Hofer2017,Sow2019}, and the Gene Ontology and Annotation (GOA) \cite{Huntley2015,Blake2013,Camon2003}. 

\subsection{Fact-Checking ChatGPT Biological Graphs}
The purpose of this step is to investigate the authenticity of contents gathered from ChatGPT and other generative AI models, and to test whether such contents may bridge the disease--genes gap in our understanding. In this regard, we propose a computational approach that captures how much true knowledge is stated in ChatGPT graphs and also identifies what may be considered noise or novelties. The idea is to compare the various link types (disease-gene, gene-gene, disease-disease) and determine how much they overlap with those in the ground truth literature graph. This offers fact checking at an aggregate level without having to verify the link semantics. Specifically, from 10 graphs constructed earlier, we implemented a process that systematically computes the number of edges in a ChatGPT-generated graph that coincide with edges in the corresponding graph derived from literature abstracts. While being in the search space, the algorithm also tracks the link to discern each type and evaluates the balance in the facts founds.  It extracts all links before it also processes one link at a time, and checks it against the ground-truth graph constructed from the literature. 

\section{Results}\label{sec2}
We used various network metrics that compare the ChatGPT graphs with literature Graphs objectively. Table \ref{tab0} encapsulates the essential metrics pertaining to each type of the knowledge graphs (i.e., literature and ChatGPT counterparts). 
\begin{table*}
\centering
\caption{The statistical result of comparing 10 GPT graphs with 10 literature graphs generated from the same number of records.}\label{tab0}
\begin{tabular}{@{}l|l|lllllllllll@{}}
\toprule
\setlength{\tabcolsep}{2pt}
% \fontsize{8pt}{8pt}Header 1 & \fontsize{10pt}{10pt}Header 2 \\ \hline
\fontsize{6pt}{6pt}Source & \fontsize{5pt}{5pt}Metric & \fontsize{4pt}{4pt}G1 & \fontsize{4pt}{4pt}G2 & \fontsize{4pt}{4pt}G3& \fontsize{4pt}{4pt}G4 & \fontsize{4pt}{4pt}G5 & \fontsize{4pt}{4pt}G6 & \fontsize{4pt}{4pt}G7 & \fontsize{4pt}{4pt}G8 & \fontsize{4pt}{4pt}G9 & G10 \\
\midrule
GPT & No of Nodes& 70 & 80 & 80 & 86 & 80 & 74 & 66 & 75 & 75 & 79 \\
PubMed & & 137 & 63 & 100 & 104 & 116 & 118 & 101 & 113 & 95 & 154 \\
% \hline
GPT & No. of Edges & 110 & 138 & 113 & 141 & 120 & 116 & 108 & 116 & 124 & 139 \\
PubMed &  & 297 & 124 & 165 & 214 & 251 & 240 & 207 & 366 & 147 & 393 \\
% \hline
GPT & N/E Ratio & 0.64 & 0.58 & 0.71 & 0.61 & 0.67 & 0.64 & 0.61 & 0.65 & 0.60 & 0.57 \\
PubMed &  & 0.46 & 0.51 & 0.61 & 0.49 & 0.46 & 0.49 & 0.49 & 0.31 & 0.65 & 0.39 \\
\midrule
GPT & No. of Diseases & 54 & 64 & 65 & 67 & 60 & 57 & 54 & 60 & 59 & 61 \\
PubMed &  & 117 & 54 & 87 & 88 & 97 & 101 & 79 & 97 & 75 & 131 \\
% \hline
GPT & No. of Genes & 16 & 16 & 15 & 19 & 20 & 17 & 12 & 15 & 16 & 18 \\
PubMed &  & 20 & 9 & 13 & 16 & 19 & 17 & 22 & 16 & 20 & 23 \\
\midrule
% \hline
GPT & Gene-Gene Link No.  & 46 & 67 & 59 & 64 & 51 & 48 & 54 & 56 & 56 & 62 \\
PubMed &  & 229 & 86 & 120 & 151 & 196 & 192 & 148 & 311 & 94 & 316 \\
GPT & Disease-Gene Link No. & 54 & 50 & 45 & 58 & 50 & 47 & 47 & 48 & 49 & 58 \\
PubMed &  & 57 & 34 & 40 & 55 & 45 & 36 & 44 & 43 & 42 & 61 \\
GPT & Disease-Disease Link No. & 10 & 19 & 9 & 19 & 19 & 20 & 6 & 11 & 19 & 19 \\
PubMed & & 11 & 4 & 5 & 5 & 10 & 10 & 14 & 10 & 11 & 16 \\
\bottomrule
\end{tabular}
\end{table*}

\begin{figure*}
\centering
% Subfigure 1 (left)
\begin{subfigure}[b]{0.45\textwidth}
    \includegraphics[width=\linewidth]{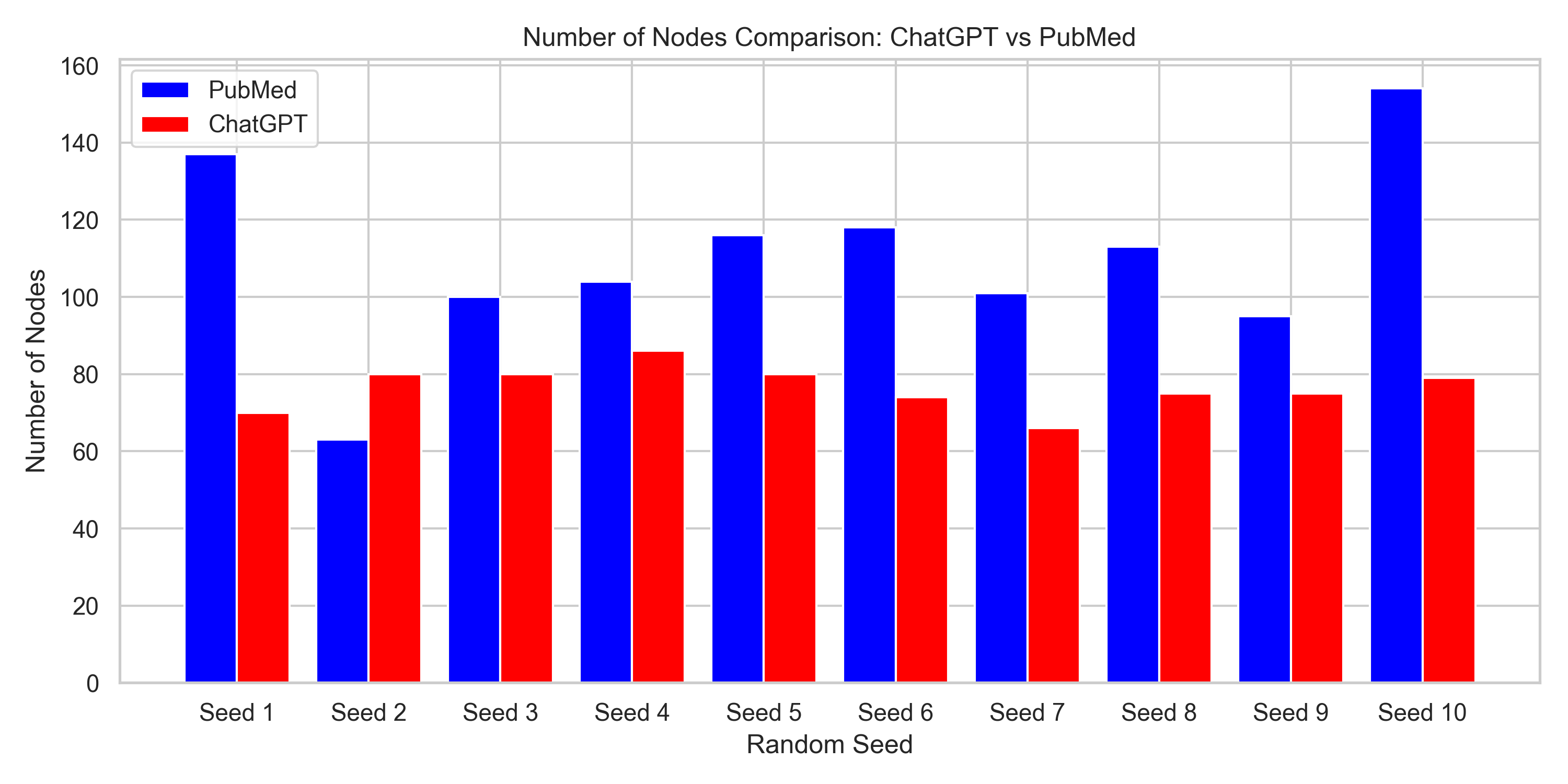}
    \caption{Nodes No.}
    \label{fig:nodes}
\end{subfigure}
% Subfigure 2 (right)
\begin{subfigure}[b]{0.45\textwidth}
    \includegraphics[width=\linewidth]{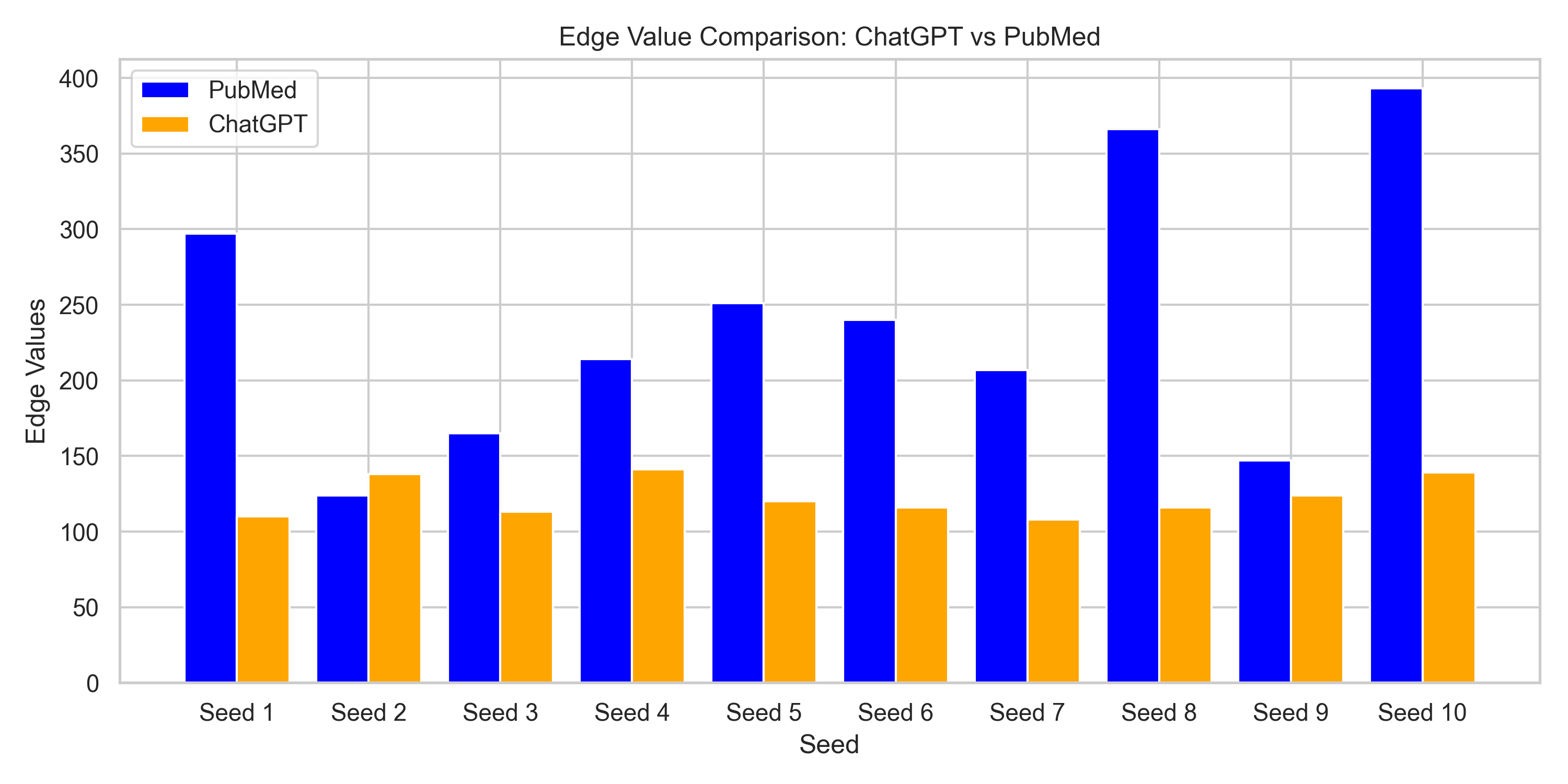}
    \caption{Edges No.}
    \label{fig:edges}
\end{subfigure}
\caption{shows two subfigures: (a) on the left, the Number of Nodes, and (b) on the right the Number of Edges comparisons of 10 chatGPT graphs against literature, respectively.}
\end{figure*}

Each consecutive two rows embody a distinct scenario for a given statistic, while the columns reference the dataset selected randomly by a given seed. The "No. of Nodes" 2-row denotes the count of all nodes, which symbolize diseases, genes. The ``No. of Edges'' 2-row unit quantifies the interconnections between nodes, reflecting relationships (e.g., disease - gene) or interactions (protein - protein). The ``N/E Ratio'' 2-row unit computes the balance between nodes and edges, potentially demonstrating the network complexity of each graph. The ``No. of Diseases'' 2-row enumerates disease-related nodes, while ``No. of Genes'' 2-row unit does the same for genes. The ``No. of Disease-Gene Links'' 2-row unit indicates associations between diseases and genes. ``No. of Disease-Disease Links'' underscores connections between different diseases. Lastly, ``Number of Gene-Gene Links'' 2-row unit unveils interactions among gene nodes. Collectively, this table provides an intricate glimpse into the network's composition, connectivity, and relationships within the biological and medical framework, fostering a deeper understanding of its underlying dynamics of each type. Figures \ref{fig:nodes} and \ref{fig:nodes} demonstrate the comparisons of nodes and edges between ChatGPT and literature, respectively.

\section{Discussion}
The discussion of our study results revolves around several key observations and findings that shed light on the comparison and potential utility of the constructed knowledge graphs. Our approach involved the comparison of two distinct types of graphs, both constructed from randomly selected datasets. This sampling strategy ensured an unbiased and fair basis for comparison between the two sources. In terms of topological analysis, it was our expectation to observe a less number of nodes and edges exhibited in ChatGPT. It was also our expectations to observe that the network generated from literature to be rich and complex, which was demonstrated by lower ratios of nodes to edges.  However, we also observed an anomalous behavior among the 10 graph. Particularly, G9 has surpassed its literature counterpart in the ratio of number of nodes to edges. Such an observation indicate complexity of certain ChatGPT graphs which warrant further pursuing.

One of the main pursuits of this work was to perform an unbiased fact-checking and verification of a truth graph constructed constructed using ontologies for their credible terminology, and biomedical literature of publications that are funded by the National Institute of Health to ensure high quality and credibility of work. We ensure that the fact-checking process is bounded by a closed-world assumption to make our work possible. The outcome of the process yielded promising results: the precision of link overlaps ranged from [70\% to 86\%] which is significantly high given the close-world assumption. This finding gives a certain measure of confidence to cautiously consider investigating data generated by ChatGPT using careful prompt-engineering.

% Though, the computational fact-checking is a core part of this work, we have taken the extra and and manually verified the gene-gene relationships identified in G1, which has the lowest fact-checking rate. We have verified that all the links existed in ChatGPT are valid links. Some of the relationships turned out to be an interaction  (e.g., SOD1, TARDBP) while others were mutations (e.g., PSEN1, PSEN2). The manual fact-checking was performed against widely acceptable databases: (1) The IntAct—open source resource for molecular interaction data \cite{kerrien2007intact}, (2) The Gene Regulatory Network Database (GRNdb) \cite{fang2021grndb,li2021single,szklarczyk2021string}. 

\section*{Conclusion and Future Direction}
As we continue to refine our work, the next steps involve further investigation of the proximity distance among biomedical terms and test if they hold in other domain and research areas. The study of disease-gene can be further instantiated in precise complex disease such as Alzheimer's and comorbidities where little is known. Such investigations may necessitate the introduction of new ontologies (e.g., Gene ontology, Drug, Chemical Entity, and drug target ontologies) among many others. In turn, this opens the door to prompt-engineer ChatGPT to answer specific questions regarding the repurposeability of a drug. Another interesting direction is to entirely \emph{retrain} the engines of ChatGPT using the confirmed-true knowledge and use its massive reasoning capabilities to answers questions about certain biological pathways to investigate a certain biological targets, or a disease that maybe caused by a certain clusters of genes. 

\section{Acknowledgments}
This research is supported by the European Union’s Horizon 2020 research and innovation programme under grant agreement Sano No 857533 and carried out within the International Research Agendas programme of the Foundation for Polish Science, co-financed by the European Union under the European Regional Development Fund, Additionally, is created as part of the Ministry of Science and Higher Education’s initiative to support the activities of Excellence Centers established in Poland under the Horizon 2020 program based on the agreement No MEiN/2023/DIR/3796'

\bibliographystyle{unsrt}   
\bibliography{references}

\end{document}